\documentclass{article}

\usepackage{arxiv}

\usepackage[utf8]{inputenc}
\usepackage[T1]{fontenc} 
\usepackage{hyperref}
\usepackage{url}

\usepackage{microtype}
\usepackage{graphicx}
\usepackage{wrapfig}
\usepackage{booktabs}

\usepackage{amsfonts, amsmath}
\usepackage{amssymb}
\usepackage{mathtools}
\usepackage{amsthm}

\usepackage{nicefrac}
\usepackage[square,sort,comma,numbers]{natbib}

\usepackage{xcolor}

\usepackage{doi}
\usepackage{calrsfs}
\usepackage{bm}
\DeclareMathAlphabet{\pazocal}{OMS}{zplm}{m}{n}
\usepackage{array}
\usepackage{multirow}
\usepackage{booktabs}

\makeatletter
\g@addto@macro{\endtabular}{\rowfont{}}% Clear row font
\makeatother
\newcommand{\rowfonttype}{}% Current row font
\newcommand{\rowfont}[1]{% Set current row font
   \gdef\rowfonttype{#1}#1%
}
\newcolumntype{L}[1]{>{\raggedright\let\newline\\\arraybackslash\rowfonttype}m{#1}}
\newcolumntype{C}[1]{>{\centering\let\newline\\\arraybackslash\rowfonttype}m{#1}}
\newcolumntype{R}[1]{>{\raggedleft\let\newline\\\arraybackslash\rowfonttype}m{#1}}
\usepackage{pgfplots}
\usetikzlibrary{arrows.meta}
\usepgfplotslibrary{groupplots}
\usepgfplotslibrary{dateplot}
\usepackage[labelfont=bf]{caption}
\usepackage{subfig}
\usepackage{float}
\definecolor{blue(pigment)}{rgb}{0.2, 0.2, 0.6}
\definecolor{darkGreen}{rgb}{0.0, 0.5, 0.0}
\definecolor{magentadark}{HTML}{C20078}

\usepackage{mfirstuc}
\MFUnocap{for}
\MFUnocap{or}
\MFUnocap{etc}
\usepackage{stfloats}
\pgfplotsset{every axis/.append style={
                    label style={font=\normalsize},
                    title style={font=\normalsize},
                    tick label style={font=\normalsize}  
                    }}

\title{\textbf{Efficient Relation-aware Neighborhood Aggregation in Graph Neural Networks via Tensor Decomposition}}

\author{{Peyman Baghershahi}\\
	University of Tehran\\
	\texttt{p.baghershahi@ut.ac.ir} \\
	\And
	{Hadi Moradi} \\
	University of Tehran\\
	\texttt{moradih@ut.ac.ir} \\
 	\And
	{Reshad Hosseini} \\
	University of Tehran\\
	\texttt{reshad.hosseini@ut.ac.ir} \\
}

\renewcommand{\shorttitle}{}

\hypersetup{
pdftitle={A template for the arxiv style},
pdfsubject={q-bio.NC, q-bio.QM},
pdfauthor={David S.~Hippocampus, Elias D.~Striatum},
pdfkeywords={First keyword, Second keyword, More},
colorlinks=true,
citecolor={blue(pigment)},
citebordercolor={blue(pigment)},
urlcolor={blue(pigment)},
}

\begin{document}
\maketitle

\begin{abstract}
Numerous Graph Neural Networks (GNNs) have been developed to tackle the challenge of Knowledge Graph Embedding (KGE). However, many of these approaches overlook the crucial role of relation information and inadequately integrate it with entity information, resulting in diminished expressive power. In this paper, we propose a novel knowledge graph encoder that incorporates tensor decomposition within the aggregation function of Relational Graph Convolutional Network (R-GCN). Our model enhances the representation of neighboring entities by employing projection matrices of a low-rank tensor defined by relation types. This approach facilitates multi-task learning, thereby generating relation-aware representations. Furthermore, we introduce a low-rank estimation technique for the core tensor through CP decomposition, which effectively compresses and regularizes our model. We adopt a training strategy inspired by contrastive learning, which relieves the training limitation of the 1-N method inherent in handling vast graphs. We outperformed all our competitors on two common benchmark datasets, FB15k-237 and WN18RR, while using low-dimensional embeddings for entities and relations. Codes are available here: \url{https://github.com/pbaghershahi/TGCN.git}.
\end{abstract}

\section{Introduction}
Knowledge Graphs (KGs) have broad applications in real-world problems. Nonetheless, their evolving nature often leads to numerous missing relations. Hence, predicting these missing relations becomes a crucial challenge known as Knowledge Graph Completion (KGC). The approach to addressing KGC involves embedding KGs in low dimensions, a process known as Knowledge Graph Embedding (KGE). These embeddings are then utilized to predict the missing links in the knowledge graph.

A group of methods within Knowledge Graph Completion (KGC) falls under the category of embedding-based approaches \cite{sun2018rotate, NIPS2013_1cecc7a7, NEURIPS2019_d961e9f2, chami-etal-2020-low}. These methods embed KGs into a feature space, preserving their semantic relations. Additionally, neural network models have demonstrated impressive performance in KGC such as Convolutional Neural Networks (CNN) \cite{dettmers2018convolutional, vashishth2020interacte, balavzevic2019hypernetwork} and Transformers \cite{wang2019:coke, BAGHERSHAHI2023110124}. However, many of these methods independently embed entities and relations without considering their local neighborhoods and the rich information within the graph structures. In contrast, Graph Neural Networks (GNNs) \cite{schlichtkrull2018modeling, Vashishth2020Composition-based, 10.1145/3442381.3450118, cucala2022explainable} are a type of method that encodes the graph structure and aggregates information over a local neighborhood. These approaches aid a new entity in acquiring an expressive representation by leveraging the information from its observed neighbors.

However, the utilization of high-dimensional embedding often raises scalability challenges for state-of-the-art methods \cite{lacroix2018canonical, NEURIPS2019_d961e9f2, chen-etal-2021-hitter} when embedding Knowledge Graphs (KGs). The problem is exacerbated when Graph Neural Networks (GNNs) generate secondary undirected graphs from the original KGs \cite{cucala2022explainable, NEURIPS2021_0fd600c9}. More importantly, many GNNs neglect the importance of relations while embedding entities due to inefficient integration of the information of entities and relations \cite{schlichtkrull2018modeling, 10.1609/aaai.v33i01.33013060, Vashishth2020Composition-based} which cannot improve expressiveness \cite{Stoica_Stretcu_2020}. On the other hand, the group of tensor decomposition-based methods \cite{balazevic-etal-2019-tucker, lacroix2018canonical, trouillon2016complex} leverage relations in encoding entities effectively, whereas they embed entities independently, neglecting the underlying graph structure in the process.

In this paper, we introduce a new framework to exploit these two paradigms to tackle their limitations. We take advantage of the Tucker decomposition in the aggregation function of R-GCN \cite{schlichtkrull2018modeling} to enhance the integration of the information of entities and relations. The Tucker decomposition offers knowledge sharing because the transformation matrices that are applied to neighboring entities are low-rank and relation-dependent which helps with efficiently extracting the interaction between an entity and the relation. Unlike previous tensor decomposition models, our method generates representations of neighboring entities instead of scoring triplets, making our model a general KG encoder. We use  CANDECOMP/PARAFAC (CP) decomposition as a regularization method for low-rank approximation of the core tensor of our model. Also, we utilize a contrastive loss that solves the scalability problem of 1-N training method for training on enormous KGs. Although utilizing low dimensionality of embeddings, our method outperforms all our competitors on both FB15k-237 and WN18RR as standard KGC benchmarks. We also show that our method improves the base R-GCN on FB15k-237 by 36\% with the same decoder.

\paragraph{Contributions:}
\begin{itemize}
    \item We propose a novel general KG encoder that employs tensor decomposition to encourage knowledge propagation. It is done by parameter sharing through relation-dependent transformations for efficient aggregation of neighborhood information. 
    \item We employ CP decomposition for low-rank approximation of the core tensor of our model to lower its number of trainable parameters and regularize it. Also, inspired by contrastive learning approaches, we train our model with a graph-size-invariant objective method to mitigate the problem of training GNNs for huge KGs. 
    \item We outperform all our powerful baselines for KGC on common benchmark datasets. 
\end{itemize}
\section{Related Work}
\paragraph{Knowledge Graph Embedding (KGE)} plays a crucial role in KGC. Embedding-based methods constitute the first category of KGC approaches is embedding-based methods wherein entities and their relations are encoded into a low-dimensional latent space to be used for prediction of potential relations based on their embedding vectors.

Translational methods \cite{NIPS2013_1cecc7a7, wang2014knowledge} involve a relation-specific translation from the source entity to the target entity, and bilinear methods \cite{DBLP:conf/icml/NickelTK11, DBLP:journals/corr/YangYHGD14a, NEURIPS2018_b2ab0019, balazevic-etal-2019-tucker} leverage tensor decomposition methods to match the latent semantics of entities. Also exploration of complex domain embeddings is particularly studied for its ability to model (anti-)symmetry and inversion relations effectively \cite{trouillon2016complex, lacroix2018canonical, sun2018rotate, NEURIPS2019_d961e9f2}. Additionally, there has been growing interest in the hyperbolic domain \cite{NEURIPS2019_f8b932c7, chami-etal-2020-low} due to its suitability for embedding hierarchical relations. 

The second category of KGC methods utilizes neural networks such as MLP \cite{socher2013reasoning}, CNN \cite{dettmers2018convolutional, vashishth2020interacte, balavzevic2019hypernetwork}, and Transformers \cite{wang2019:coke, chen-etal-2021-hitter, BAGHERSHAHI2023110124}. Unlike embedding-based methods, neural network approaches adopt free parameters to share learned knowledge throughout the entire KG. However, these methods should adapt to evolving nature of KGs and generalize to unseen data in inductive settings.

\paragraph{Graph Neural Networks (GNNs)} have found extensive application across recommendation systems, 3d geometric, bioinformatics, traffic control, and natural language processing \cite{10.1145/3459637.3481916, 10.1145/3326362, 10.1093/bioinformatics/bty294}. These models encode graph entities into a low-dimensional feature space, taking into account their contextual relationships. The resulting representations are thus enriched with the inherent structure of the graph. Most GNNs deploy convolution operations on local neighborhood of fixed size, employing the adjacency matrix of the whole graph in the spectral domain \cite{kipf2017semisupervised, pmlr-v97-wu19e}, or of varying size by sampling in the spatial domain \cite{veličković2018graph, NIPS2017_5dd9db5e}. Moreover, some methods encode entities based on their relative position in the graph \cite{https://doi.org/10.48550/arxiv.1906.04817}. 

\paragraph{GNNs in KGE} Over the past few years, GNNs have become prevalent in KGE tasks. Our approach shares similarities with R-GCN \cite{schlichtkrull2018modeling}, while it assigns projection matrices to each relation type, however, solely utilizes relation embedding vectors for triplet scoring, failing to adequately capture relation-entity interactions. Moreover, incorporating new relations necessitates additional projection matrices, increasing the risk of overfitting. SACN \cite{10.1609/aaai.v33i01.33013060} adopts a weighted GCN to aggregate varying degrees of information from neighboring entities. On the other hand, COMPGCN \cite{Vashishth2020Composition-based} restricts relation-specific projections to input, output, and self-loops, updating both entity and relation representations. Nonetheless, COMPGCN similarly struggles with interaction capture and relies on a fixed adjacency matrix, limiting its applicability to unseen entities.

Moving towards the burgeoning interest in hyperbolic space embedding of KGs, M${}^2$GNN \cite{10.1145/3442381.3450118} mixes multiple single-curvature spaces into a unified space, aligning with the heterogeneity of KG data. Leveraging a GNN, it exploits the local neighborhood of triplets. Similarly, ConE \cite{NEURIPS2021_662a2e96} introduces a model that embeds a KG into the product space of multiple hyperbolic planes, corresponding to 2D hyperbolic cones, to effectively model both hierarchical and non-hierarchical relations.

The attention mechanism plays a crucial role in the realm of GNNs designed for KGC. Models like A2N \cite{bansal2019a2n}, CoKE \cite{wang2019:coke}, SAttLE \cite{BAGHERSHAHI2023110124}, and HittER \cite{chen-etal-2021-hitter} incorporate attention to effectively weight information from neighboring nodes and generate query-specific representations. To capture long-range dependencies while avoiding over-smoothing, MAGNA \cite{ijcai2021p0425} employs a strategy of diffusing attention across both directly and indirectly connected entities within a specified multi-hop neighborhood.

KGs are intrinsically undergo evolution, with common addition of unseen entities and relations. GNNs typically are more inductive than other models methods for KGE \cite{10.5555/3172077.3172138} because they encode entities based on their local neighborhoods, however, there is a notable interest in enhancing their inductive capabilities. Regarding this, GraIL \cite{10.5555/3524938.3525814} introduces a method where predictions are detached from entity-specific embeddings by labeling entities based on enclosing subgraphs and scoring them using a GNN akin to RGCN. INDIGO \cite{NEURIPS2021_0fd600c9} eliminates reliance on predefined scoring functions by generating an undirected annotated secondary graph and decoding entity representations from the outermost layer of a GCN. Additionally, MGNN \cite{cucala2022explainable} learns transformations to make predictions explainable through sets of rules. 
\section{Background}
\paragraph{Problem} Given a knowledge graph $\pazocal{KG}(V, R)$ consisted of triplets $(s, r, t)$, with $s, t \in \pazocal{V}$ being the source and target entities. These entities (graph nodes) are connected with relation type $r \in \pazocal{R}$, where $\pazocal{V}$ and $\pazocal{R}$ are the sets of all types of entities and relations. The Knowledge Graph Completion (KGC) task is to verify how probable an unseen triplet $(s, r, t)$ truly exists in $\pazocal{KG}$.

\textbf{R-GCN:}
R-GCN generalizes GCN \cite{kipf2017semisupervised} to relational data. This model projects the representations of the neighboring entities by weight matrices indexed by their relation types. It projects the latent embedding of the entity by a loop weight matrix. Then it computes a normalized aggregation over all projected representations. The propagation model of R-GCN for an entity $v$ is as follows:
\vspace{-0.02cm}
\begin{align}\label{eq:rgcn}
	h_{v}^{(l+1)}=\sigma(\sum_{r\in \pazocal{R}}\sum_{u\in \pazocal{N}_{v}^{r}}\frac{1}{c_{v,r}}W_r^{(l)}h_u^{(l)}+W_0^{(l)}h_v^{(l)})
\end{align}
\vspace{-0.02cm}
in which $l$ is the l-th layer of the model, $\pazocal{N}_{v}^{r}$ is the set of neighbors of $v$, $r\in \pazocal{R}$ is the relation type of each neighbor, $c_{v,r}$ is a normalization factor, $W_r \in \mathbb{R}^{d \times d}$ is the weight matrix assigned to $r$, and $W_0 \in \mathbb{R}^{d \times d}$ is a loop weight matrix and $d$ is the dimensionality of embeddings. Also, to prevent overfitting due to the increasing number of relations, R-GCN proposes two regularization methods.

\textbf{TuckER:}
TuckER \cite{balazevic-etal-2019-tucker} is a linear model which uses Tucker decomposition \cite{tucker1966some}. The low-rank core tensor in TcukER offers multi-task learning by parameter sharing through relations. In contrast to R-GCN, TuckER is not a KG encoder and does not include graph structure information. TuckER scores triplets directly using as follows:
\vspace{-0.02cm}
\begin{align}\label{eq:tucker}
	\phi(s, r, t)=W_c\times_1 e_s \times_2 e_r \times_3 e_t
\end{align}
\vspace{-0.02cm}
in which \(W_c \in \mathbb{R}^{d_e \times d_r \times d_e}\) is the learnable core tensor, $e_s, e_t \in \mathbb{R}^{d_e}$ are the embeddings of $s$ and $t$ entities with dimensionality $d_e$, and $e_r \in \mathbb{R}^{d_r}$ is the embedding of $r$ with dimensionality $d_r$. $\times_n$ indicates n-mode tensor product.
\section{Methods}
In KGs, relations have as rich information as entities. Although R-GCN generalizes GCN to KGs, its weakness is inefficiently combining the information of relations and entities. In other words, it uses relations for indexing weight matrices that project representations of entities. This method cannot enhance the expressiveness of the model \cite{Stoica_Stretcu_2020} because the learned knowledge of entities is not shared with relations. Therefore, R-GCN does not optimize the embedding of relations properly. 

In this work, we propose \textit{\textbf{T}ucker \textbf{G}raph \textbf{C}onvolutional \textbf{N}etworks (\textbf{TGCN})}, a GNN model to address the above limitations. Inspired by TuckER, we change the aggregation function of R-GCN to take advantage of parameter sharing by multi-task learning through relations. In our proposed aggregation function, the representations of the neighbors are transformed by applying learnable weight matrices, which their parameters are defined by the embedding of the relation type of each entity. Following R-GCN, a universal projection matrix is applied to the representation of the entity itself. The output of each TGCN layer for an entity is computed as a normalized sum of all projected representations. Our propagation model for updating the representation of an entity is as follows:
\vspace{-0.02cm}
\begin{align}\label{eq:tgcn}
	h_{v}^{(l+1)}=\sigma(\sum_{r\in \pazocal{R}}\sum_{u\in \pazocal{N}_{v}^{r}}\frac{1}{c_{v,r}}f(e_r, h_u^{(l)})+W_0^{(l)}h_v^{(l)})
\end{align}
\vspace{-0.02cm}
where $f(e_r, h_u^{(l)})=W_c^{(l)}\times_1 h_u^{(l)} \times_2 e_r$ is a function of each neighboring entity representation and its relation type embedding. Here, $\pazocal{N}_{v}^{r}$ is the set of neighbors of $v$ are related by $r\in \pazocal{R}$, $c_{v,r}$ is a normalization factor, $e_r \in \mathbb{R}^{d_r}$ is the embedding of $r$, $W_c^{(l)} \in \mathbb{R}^{d_e\times d_r \times d_e}$ is the core weight tensor of layer $l$, and $W_0 \in \mathbb{R}^{d_e \times d_e}$ is the loop weight matrix. Also, each $v$ is assigned to an embedding vector $e_v \in \mathbb{R}^{d_e}$ and $h_{v}^{0} = e_v$. $d_e$ and $d_r$ are the dimensionalities of the embeddings of entities and relations respectively.

Equation \ref{eq:tgcn} shows that contrary to TuckER, TGCN embeds the rich information of the graph structure in its representations. Also, learned knowledge is accessible through the whole KG by parameter sharing due to the low-rank core tensor. 

\subsection{Model Compression}
The number of trainable parameters of the Tucker core tensor is of $O(d_e^2d_r)$. Therefore, in case of using high dimensional embeddings, this leads to overfitting and memory issues. To tackle this problem, we take a step forward in terms of parameter sharing and introduce a model compression method utilizing CANDECOMP/PARAFAC (CP) decomposition \cite{https://doi.org/10.1002/1099-128X(200005/06)14:3<105::AID-CEM582>3.0.CO;2-I, hitchcock1927expression} for low-rank approximation of the core tensor. This approximation encourages multi-task learning more and significantly decreases the number of free parameters of the core tensor. We also anticipate that the method can be used as an effective regularization method.

\textbf{CP Decomposition:} As a tensor rank decomposition method, CP decomposition factorizes a tensor into a sum of rank-one tensors. An approximation of a third-order tensor $\pazocal{X} \in \mathbb{R}^{I \times J \times K}$ using CP decomposition is as follows:
\vspace{-0.02cm}
\begin{align}\label{eq:cp3}
	\pazocal{X} \approx \sum_{r=1}^{R} a_r \circ b_r \circ c_r
\end{align}
\vspace{-0.02cm}
in which $R$ is the rank of the tensor and $a_r \in \mathbb{R}^I$, $b_r \in \mathbb{R}^J$, and $c_r \in \mathbb{R}^K$.

If we define $A=[a_1\ a_2\ \cdots a_R]$,  $B = [b_1\ b_2\ \cdots b_R]$ and $C = [c_1\ c_2\ \cdots c_R]$ as factor matrices of the rank-one components, then the above model can be written in terms of the frontal slices of $\pazocal{X}$:
\vspace{-0.02cm}
\begin{align}\label{eq:cp5}
	X_k \approx AD^{(k)}B^T, \quad \text{in which} \quad D^{(k)}\equiv
	diag(c_{k:}) \quad \text{for} \quad k=1,\cdots,K
\end{align}
\vspace{-0.02cm}
in which $A$ is a factor matrix of the rank-one components. Finally, we can write CP decomposition in short notation as:
\vspace{-0.02cm}
\begin{align}\label{eq:cp6}
	\pazocal{X} \approx [[A, B, C]] \equiv \sum_{r=1}^{R} a_r \circ b_r \circ c_r
\end{align}
\vspace{-0.02cm}
Now, we rewrite the TGCN propagation model for updating an entity representation using CP decomposition as follows:
\vspace{-0.02cm}
\begin{align}\label{eq:tgcnCP}
	h_{v}^{(l+1)}=\sigma(\sum_{r\in \pazocal{R}}&\sum_{u\in \pazocal{N}_{v}^{r}}\frac{1}{c_{v,r}}f(e_r, h_u^{(l)})+W_0^{(l)}h_v^{(l)}) \\
	\text{s.t.}\quad f(e_r, h_u^{(l)})&=W_c^{(l)}\times_1 h_u^{(l)} \times_2 e_r \notag\\
	W_c^{(l)} &= [[W_1^{(l)}W_2^{(l)}W_3^{(l)}]] \notag
\end{align}
\vspace{-0.02cm}
in which $W_c^{(l)} \in \mathbb{R}^{d_e\times d_r \times d_e}$ is a product of tensors $W_1^{(l)}\in \mathbb{R}^{n_b \times d_r}$, $W_2^{(l)}\in \mathbb{R}^{n_b \times d_e}$, and $W_3^{(l)}\in \mathbb{R}^{n_b \times d_o}$. Finding matrices $W_1^{(l)}$, $W_2^{(l)}$, and $W_3^{(l)}$ is originally an optimization problem, but in our case, we try to learn these matrices. Therefore, we select a value for $n_b$ which we call the number of bases and it is equivalent to the rank ($R$) of the approximated tensor.

Our general KG encoder can integrate with many decoding methods (scoring functions). Here, we use DistMult \cite{DBLP:journals/corr/YangYHGD14a} and TuckER \cite{balazevic-etal-2019-tucker} and consider $h_s, h_t  \in \mathbb{R}^{d_e}$  being the generated representations for source and target entities.

DistMult is a fast and simple decoder without extra parameters and which computes a triplet score by a three-way multiplication as follows:
\vspace{-0.02cm}
\begin{align}\label{eq:tgcnDistmult}
	\phi(s, r, t)= h_s^T \times_2 E_r \times_3 h_t, \quad \text{where} \quad E_r \equiv diag(e_r)
\end{align}
\vspace{-0.02cm}
Previously we used Tucker decomposition to produce representation vectors. However, it can be used to score triplets in the following way as proposed in TuckER:
\vspace{-0.02cm}
\begin{align}\label{eq:tgcnTucker}
	\phi(s, r, t) = W_c \times_1 h_s \times_2 e_r \times_3 h_t
\end{align}
\vspace{-0.02cm}
Eventually, we add a logistic sigmoid layer to evaluate how probable a triplet $(s, r, t)$ is. Notably, we expect TGCN to perform better using more efficient decoding methods.

\subsection{Training}
Most state-of-the-art models follow 1-N method \cite{dettmers2018convolutional} for training, and they use Binary Cross Entropy (BCE) loss function for optimization. For a triplet $(s, r, t)$ the loss function is defined as:
\vspace{-0.02cm}
\begin{align}
	\label{eq:tgcnBCELoss}
	\pazocal{L}=-\frac{1}{|\pazocal{V}|}\sum_{t' \in \pazocal{V}} y_{t'}\log{p_{t'}}&+(1-y_{t'})\log{(1-p_{t'})}\\
	\nonumber\text{where }p_{t'}&=\sigma(\phi(s,r,t')),\\
	\nonumber\text{and }y_{t'}&= \left\{ \begin{array}{rcl}
		1 & \text{if } t'=t \\
		0 & \text{otherwise}
	\end{array}\right.
\end{align}
\vspace{-0.02cm}
in which $\sigma(.)$ is the logistic sigmoid function.

Although the 1-N approach has good performance, for each training iteration, it requires $O(|\pazocal{V}|^2d)$ operations. Thus, a model faces scalability issues while $|\pazocal{V}|$ could be a large number.

Since KGC is a self-supervised problem, we approach the above issue by a training method inspired by contrastive learning which aids in producing expressive representations by preserving high-level features while filtering out low-level features, such as noise \cite{https://doi.org/10.48550/arxiv.1807.03748}. Specifically, our training method is based on SimCLR \cite{10.5555/3524938.3525087}.

\textbf{SimCLR:}
A contrastive approach is in which two augmented samples are generated from each image in a training batch. The objective is to make the model-generated representations of these samples close to each other and distant from other samples within the batch. SimCLR incorporates the NT-Xent loss for training.

\subsubsection{1-b Training}
We are motivated by the fact that in KGs, two entities with a relation are likely to be close to each other in feature space compared to entities to which they are not connected. This intuition is similar to that of SimCLR. Therefore, we use NT-Xent loss as follows:
\vspace{-0.02cm}
\begin{align}\label{eq:1Bloss}
	\pazocal{L}=-\log(\frac{\exp(\phi(s, r, t)/\tau)}{\sum_{t'\in \pazocal{E}}\exp(\phi(s, r, t')\tau)})
\end{align}
\vspace{-0.02cm}
in which $\tau$ is temperature, and $\pazocal{E}$ is the set of entities in each training batch.

Contrary to 1-N training method, multiplication operations in 1-b method are just between unique entities of a single batch. Hence, the operational complexity of each iteration reduces to $O(|E|^2d)$ and $|E|$ can be set based on the computational limitations. We anticipate the representation generated by NT-Xent loss is more expressive than BCE loss, but we leave this for future research.
\section{Experiments}

\subsection{Using Random Subgraphs}
\label{sec:subgraph}
To train TGCN we randomly take subgraphs from the whole KG in each iteration. This is necessary because real-world KGs are typically enormous, making it impractical to process them in their entirety. Utilizing random subgraphs during training is akin to injecting noise, thereby introducing a regularization effect that enhances performance. However, maintaining an appropriate subgraph size is crucial. On one hand, a significantly large random subgraph introduces excessive noise, particularly when entities in large KGs have high degree centrality, leading to counterproductive noise. On the other hand, if we keep the size of random subgraphs constant, each entity can still access more distant neighbors and their information. As a result, the size of random subgraphs, denoted as $g_s$, plays a pivotal role in shaping the performance of our model.

\subsection{Datasets}
FB15k-237 \citep{toutanova2015representing} and WN18RR \citep{dettmers2018convolutional} are the standard KGC benchmarks we used for the evaluation of our model \cite{dettmers2018convolutional}. Find the statistics of the datasets in the Appendix \ref{appendix:datasets}. For each dataset, we add $(t, r^{-1}, s)$ as the inverse of each triplet $(s, r, t)$, known as reciprocal learning \citep{lacroix2018canonical}, with $r^{-1}$ being the inverse of $r$. The model is trained using both original triplets and their inverse ones.

\subsection{Evaluation Protocol}
\label{subappendix:evalProtocol}
We used Mean Reciprocal Rank (MRR) and Hits@\(k\) ratios in \textit{filtered} setting \cite{NIPS2013_1cecc7a7} as two commonly used ranking-based metrics for KGC to to evaluate our model. For fair evaluation, the random protocol proposed by \cite{sun-etal-2020-evaluation} is used to randomly shuffle all triplets before scoring and sorting them.

\section{Results}
We evaluate the performance of TGCN against a few of the best GNN-based encoders and a few powerful baseline models. Our experimental results on FB15k-237 and WN18RR are illustrated in Table \ref{tab:tgcnComp}. Overall, TGCN achieves the best results on both datasets. Worthwhile to note, most state-of-the-art models and our competitors, like Tucker, use high-dimensional embeddings, leading to severe scalability issues on massive KGs in web or social media applications. However, TGCN employs a comparatively lower dimensionality of embedding, alleviating this problem.

\begin{table}[t!]
    \caption{KGC results on the benchmark datasets. Results of $\dagger$ are taken from \citep{dettmers2018convolutional} and results of $\ddagger$ are taken from \citep{sun2018rotate}. Other results are taken from original papers. DoE stands for Dimensionality of Embeddings. Overall, TGCN-Tucker has superior performance to all baselines on both datasets although it has the lowest dimensionality of embeddings. TGCN-Distmult improves the simple corresponding decoder DistMult significantly.}
	\vskip 0.1in
	% \footnotesize
	\centerline{
			\begin{tabular}{L{1.1in}C{0.2in}*{8}{C{0.35in}}}
				\toprule
				\multirow{2}{4em}{} & \multirow{2}{4em}{\textbf{DoE}} & \multicolumn{4}{c}{\textbf{FB15k-237}} &
				\multicolumn{4}{c}{\textbf{WN18RR}} \\
				\cmidrule(lr){3-6} \cmidrule(lr){7-10}
				&  &   MRR   &   Hits@1   &   Hits@3   &   Hits@10   &   MRR   &   Hits@1   &   Hits@3   &   Hits@10 \\
				\midrule
				% \rowfont{\footnotesize}
				DistMult$\dagger$ \citep{schlichtkrull2018modeling}   &   100   &   .241   &   .155   &   .263   &   .419   &   .430   &   .390   &   .440   &   .490 \\
				R-GCN \citep{schlichtkrull2018modeling}   &   500   &   .250   &   .150   &   .260   &   .420   &   ---   &   ---   &   ---   &   --- \\
				ComplEx$\ddagger$ \citep{trouillon2016complex}  &    400   &   .247   &   .158   &   .275   &   .428   &   .440   &   .410   &   .460   &   .51 \\
				ConvE \citep{dettmers2018convolutional}   &   200   &   .325   &   .237   &   .356   &   .501   &   .430   &   .400   &   .440   &   .520 \\
				RotatE \citep{sun2018rotate}   &   1000   &   .338   &   .241   &   .375   &   .533   &   .476   &   .428   &   .492   &   \textbf{.571} \\
				SACN \citep{10.1609/aaai.v33i01.33013060}   &   200   &   .350   &   .260   &   .390   &   .540   &   .470   &   .430   &   .480   &   .540 \\
				COMPGCN \citep{Vashishth2020Composition-based}   &   100   &   .355   &   .264   &   .390   &   .535   &   \underline{.479}   &   \underline{.443}   &   \underline{.494}   &   .546 \\
				TuckER \citep{balazevic-etal-2019-tucker}   &   200   &   \underline{.358}   &   \textbf{.266}   &   \underline{.394}   &   \textbf{.544}   &   .470   &   \textbf{.444}   &   .482   &   .526 \\
				\midrule
				TGCN-DistMult   &   \textbf{100}   &   .339   &   .249   &   .370   &   .517   &   .452   &   .419   &   .461   &   .516 \\
				TGCN-Tucker   &   \textbf{100}   &   \textbf{.359}   &  \textbf{.266}   &   \textbf{.396}   &   \underline{.542}   &   \textbf{.482}   &   .441   &   \textbf{.500}   &   \underline{.560} \\
				\bottomrule
			\end{tabular}
			}
	\label{tab:tgcnComp}
	% 	\vskip -0.1in
\end{table}

On FB15k-237, TuckER is our closets competitor. It takes advantage of multi-task learning by parameter sharing through relations. Formally, relation types define the projection matrices applying to the representations of source entities, so entities having the same relation are projected into the same region of a feature space. This approach, while effective, reveals challenges in WN18RR with fewer relation types, causing overlapped feature space regions with indistinguishable decision boundaries between. 

On the contrary, the projection matrices of TGCN encoder are applied to the neighbors of entities instead of the entities directly, and their projected representations are aggregated and combined with the previous information of the entities. Therefore, TGCN alleviates the above problem and it performs well on both datasets. 

Lastly, unlike decoding/scoring methods such as TuckER, which are specifically proposed for link prediction, TGCN, COMPGCN, SACN, and R-GCN are general KG encoders so are applicable in node-level and graph-level tasks as well. Also, these encoders can integrate with different decoding methods for KGC.

\subsection{The Effect of The Decoding Methods}
Table \ref{tab:tgcnComp} clearly shows that TGCN can adaptively combine with decoding methods to improve their performance. TGCN has improved the MRR of DistMult by $40.7\%$ on FB15k-237 and by $5\%$ on WN18RR. Moreover, in combination with TuckER decoder, TGCN has increased its MRR by $2\%$ on WN18RR and by $0.1\%$ on FB15k-237. Though, it is noteworthy that TGCN decreases DoE by $100\%$ compared to TuckER.

\begin{table}[t!]
    \caption{
    Effect of CP decomposition as model compression method on TGCN performance and the number of free parameters of the model. This method considerably reduces the number of trainable parameters on both datasets with different decoders.
    }
	\vskip 0.1in
	\footnotesize
	\centerline{
			\begin{tabular}{L{0.7in}C{0.5in}C{0.2in}*{6}{C{0.35in}}}
				\toprule
				& \textbf{Decoder} &  \textbf{CP} &   MRR   &   Hits@1   &   Hits@3   &   Hits@10   &   \#NFP   &   \#EFP  \\
				\midrule
				\multirow{4}{6em}{\textbf{FB15k-237}}   &  \multirow{2}{4em}{DistMult}  &   No   &   .339   &   .249   &   .370   &   .517   &   2.02M   &   \multirow{4}{4em}{1.50M}  \\
				&  &   Yes   &   .334   &  .246   &   .365   &   .510   &   0.08M   &   \\
				\cmidrule(lr){2-8}
				&  \multirow{2}{4em}{TuckER}  &   No   &   .359   &  .266   &   .396   &   .542   &   3.02M   &   \\
				&  &   Yes   &   .353   &  .263   &   .387   &   .532   &   1.08M   &  \\
				\midrule
				\multirow{4}{4em}{\textbf{WN18RR}}   &  \multirow{2}{4em}{DistMult}  &   No   &   .452   &   .419   &   .461   &   .516   &   2.02M   &   \multirow{4}{4em}{4.10M}   \\
				&  &   Yes   &   .437   &  .264   &   .392   &   .535   &   0.08M   &   \\
				\cmidrule(lr){2-8}
				&  \multirow{2}{4em}{TuckER}  &   No   &   .482   &   .441   &   .500   &   .560   &   3.02M   &   \\
				&  &   Yes   &   .471   &   .438   &   .484   &   .532   &   1.08M   &   \\
				\bottomrule
			\end{tabular}
			}
	\label{tab:cpComp}
	% 	\vskip -0.1in
\end{table} 

\subsection{The Effect of The Model Compression Method}
Our empirical results on FB15k-237 and WN18RR with and without using CP decomposition for model compression are demonstrated in Table \ref{tab:cpComp}. The number of Embedding Free Parameters (\#EFP) indicates the number of free parameters of the embedding matrices for entities and relations. The number of Nonembedding Free Parameters (\#NFP) indicates the total number of all other free parameters of the model. We show more experimental results of the effectiveness of using CP decomposition for regularization in the Appendix \ref{appendix:experiments}.

Using low-rank decomposition of the core tensor by CP decomposition has significantly decreased the free parameters of TGCN encoder on both datasets. Thus they can be ignored and only the parameters of the decoder remain.

The performance gap between using CP decomposition and not using it on FB15k-237 is minor, whereas it is more evident on WN18RR. We conjecture that the decrease in performance when employing CP decomposition might be attributed to excessive parameter sharing through relations. The diversity of relations in WN18RR is considerably lower than in FB15k-237. Therefore, when making a low-rank approximation of the core tensor, the transformations applied to neighbors become more similar, resulting in overlapping projection spaces. A higher number of relations alleviates this overlap, and the results on FB15k-237 indicate that the performance difference is not considerable.

It can be seen in Table \ref{tab:cpComp} that in the case of using DisMult Decoder, \#NFP is lower than 0.1M. Importantly, TGCN has improved DistMult performance on FB15k-237 and WN18RR. This considerable performance increase is attained by a negligible number of extra parameters, which proves the effectiveness of TGCN as a general encoder.

\begin{figure*}[b!]
    \centering
    \vskip 0.1in
    \begin{tikzpicture}
                \begin{axis}[
                    width=2.5in,
                    xlabel={$n_b$},
                    xlabel near ticks,
                    ylabel={MRR},
                    ylabel near ticks,
                    % xlabel style={
                    %     yshift = {0}, font=\small
                    % },
                    % ylabel style={
                    %     yshift = {10}, font=\small
                    % },
                    ylabel style={font=\small},
                    xmin=45, xmax=2100,
                    ymin=0.46, ymax=0.472,
                    axis y line*=left,
                    xtick={50,100,250,500,1000,2000},
                    ytick={0.461, 0.470, 0.471, 0.472},
                    xmode=log,
                    %				log basis x={10},
                    xticklabel style={font=\footnotesize},
                    xticklabel={\pgfmathparse{ceil(exp(\tick))}\pgfmathprintnumber{\pgfmathresult}},
                    yticklabel style = {
                        font=\footnotesize,
                        scaled ticks=false,
                        /pgf/number format/fixed,
                        /pgf/number format/precision=3
                    },
                    % yticklabel style = {
                    %     font=\footnotesize,
                    %     /pgf/number format/.cd,
                    %     fixed,
                    %     fixed zerofill,
                    %     precision=3,
                    %     /tikz/.cd
                    % },
                    yticklabel={\pgfmathprintnumber{\tick}},
                    % legend pos=south east,
                    ymajorgrids=true,
                    grid style=dashed,
                    ]
                    \addplot[color=darkGreen, mark=square, mark options={scale=0.75}, line width=1pt]
                    coordinates {
                        (50, 0.461)
                        (100, 0.471)
                        (250, 0.471)
                        (500, 0.471)
                        (1000, 0.472)
                        (2000, 0.472)
                    };
                    % \addlegendentry{TGCN-Tucker}
                \end{axis}
                \label{fig:MRRvsnb}
                \begin{axis}[
                    width=2.5in,
                    ylabel={\#ENFP},
                    ylabel near ticks,
                    % xlabel style={
                    %     yshift = {0}, font=\small
                    % },
                    % ylabel style={
                    %     yshift = {10}, font=\small
                    % },
                    ylabel style={font=\small},
                    xmin=45, xmax=2100,
                    ymin=0.0, ymax=1.3,
                    hide x axis,
                    axis y line*=right,
                    ytick={0.05, 0.171, 0.321, 0.621, 1.221},
                    xmode=log,
                    xticklabel={\pgfmathparse{ceil(exp(\tick))}\pgfmathprintnumber{\pgfmathresult}},
                    yticklabel style = {
                        font=\footnotesize,
                        scaled ticks=false,
                        /pgf/number format/fixed,
                        /pgf/number format/precision=2
                    },
                    yticklabel={\pgfmathprintnumber{\tick}M},
                    % legend pos=south east,
                    ymajorgrids=true,
                    grid style=dashed,
                    ]
                    \addplot[color=cyan, mark=square, mark options={scale=0.75}, line width=1pt, line width=1pt]
                    coordinates {
                        (50, 0.051)
                        (100, 0.081)
                        (250, 0.171)
                        (500, 0.321)
                        (1000, 0.621)
                        (2000, 1.221)
                    };
                    % \addlegendentry{TGCN-Tucker}
                \end{axis}
                \label{fig:nbvsenfp}
            \end{tikzpicture}
            \vspace{0.05in}
            \caption{
                The effect of the number of bases ($n_b$) for CP decomposition. The green line shows the performance of TGCN-Tucker on WN18RR for $d=100$ with different $n_b$ and the blue line shows its \#ENFP. Both figures are on a logarithmic scale, meaning that \#ENFP increases linearly by increasing $n_b$ although its curve it if exponential.
            }
    \label{fig:nb}
    \vskip -0.1in
\end{figure*}
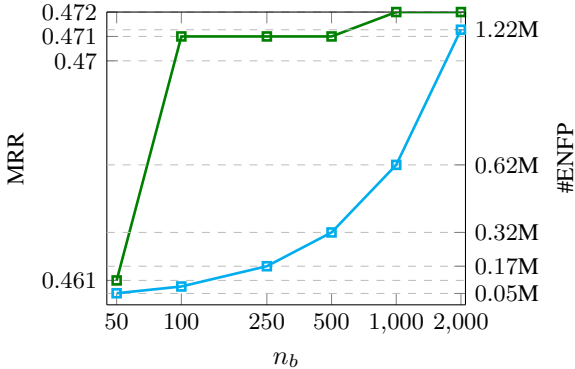

\subsubsection{The Effect of The Number of Bases}
To investigate the effect of the number of bases on TGCN using CP decomposition, we show model performance and the number of Encoder Nonembedding Free Parameters (\#ENFP) as a function of the number of bases $n_b$. Figure \ref{fig:nb} shows the results of this experiment. The Hyper-parameters are only tuned for $n_b=100$, and fixed for other values of $n_b$.

As we expected, the number of bases is effective since the performance of TGCN obviously increases slightly using a high number of bases. On the other hand, its \#ENFP increases linearly, leading to more complexity. Besides, we can see that \#ENFP is comparably low even in high values of $n_b$, and we can neglect it in $n_b=100$. So we choose $n_b=100$ for a trade-off. This shows that TGCN is highly applicable for low memory and computation limitations.

\begin{figure}[t!]
	\centering
	\vskip 0.1in
	\subfloat[MRR vs $g_s$ on FB15k-237]{
			\begin{tikzpicture}
				\begin{axis}[
					width=2.5in,
					xlabel={$g_s$},
                        ylabel={MRR},
                        xlabel near ticks,
                        ylabel near ticks,
					% xlabel style={
					% 	yshift = {0}, font=\small
					% },
					% ylabel style={
					% 	yshift = {10}, font=\small
					% },
                        ylabel style={font=\small},
					xmin=8000, xmax=102000,
					ymin=0.325, ymax=0.362,
					xtick={10000,20000,30000,40000,50000,60000,70000,80000,90000, 100000},
					ytick={0.330, 0.340, 0.345, 0.35, 0.355, 0.359},
					xticklabel style={font=\footnotesize},
					xticklabel={\pgfmathprintnumber{\tick}},
					yticklabel style = {
						font=\footnotesize,
						/pgf/number format/.cd,
						fixed,
						fixed zerofill,
						precision=3,
						/tikz/.cd
					},
					yticklabel={\pgfmathprintnumber{\tick}},
					legend pos=south east,
					ymajorgrids=true,
					grid style=dashed,
					legend style={font=\scriptsize},
					]
					\addplot[color=magenta, mark=square, mark options={scale=0.75}, line width=1pt]
					coordinates {
						(10000, 0.330)
						(20000, 0.340)
						(30000, 0.345)
						(40000, 0.348)
						(50000, 0.349)
						(60000, 0.351)
						(70000, 0.353)
						(80000, 0.356)
						(90000, 0.358)
                            (100000, 0.359)
					};
					% \addlegendentry{TGCN-Tucker}
				\end{axis}
				\label{fig:fbMRRvsgs}
			\end{tikzpicture}
	}%
	\qquad
	\subfloat[MRR vs $g_s$ on WN18RR]{
			\begin{tikzpicture}
				\begin{axis}[
					width=2.5in,
					xlabel={$g_s$},
                        ylabel={MRR},
                        xlabel near ticks,
                        ylabel near ticks,
					% xlabel style={
					% 	yshift = {0}, font=\small
					% },
					% ylabel style={
					% 	yshift = {10}, font=\small
					% },
                        ylabel style={font=\small},
					xmin=8000, xmax=72000,
					ymin=0.444, ymax=0.486,
					xtick={10000,20000,30000,40000,50000,60000,70000},
					ytick={0.445, 0.460, 0.465, 0.470, 0.475, 0.480, 0.484},
					xticklabel style={font=\footnotesize},
					xticklabel={\pgfmathprintnumber{\tick}},
					yticklabel style = {
						font=\footnotesize,
						/pgf/number format/.cd,
						fixed,
						fixed zerofill,
						precision=3,
						/tikz/.cd
					},
					yticklabel={\pgfmathprintnumber{\tick}},
					legend pos=south east,
					ymajorgrids=true,
					grid style=dashed,
					legend style={font=\scriptsize},
					]
					\addplot[color=red, mark=square, mark options={scale=0.75}, line width=1pt]
					coordinates {
						(10000, 0.449)
						(20000, 0.465)
						(30000, 0.476)
						(40000, 0.479)
						(50000, 0.484)
						(60000, 0.483)
						(70000, 0.481)
					};
					% \addlegendentry{TGCN-Tucker}
				\end{axis}
				\label{fig:wnMRRvsgs}
			\end{tikzpicture}
	}%
	\caption{
	The effect of the size of the random subgraphs ($g_s$) for training on the performance of TGCN-Tucker on FB15k-237 and WN188.
	}
	\label{fig:gs}
	\vskip -0.1in
\end{figure}
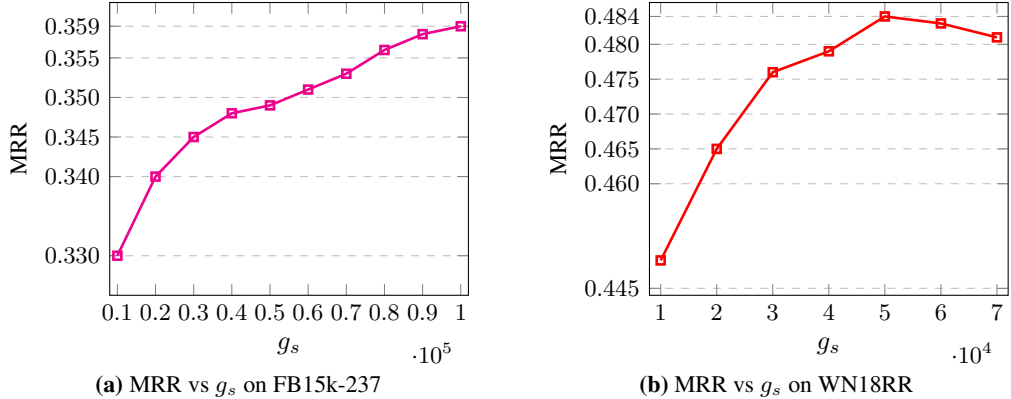

\begin{table}[t!]
    \caption{The effect of the dimensionality of embeddings. Results show that TGCN has competitive performance even with low-dimensional embeddings making it highly scalable to huge datasets because of the reduction in required memory and computation.}
	\vskip 0.1in
	% \footnotesize
	\centerline{
			\begin{tabular}{L{0.8in}C{0.2in}*{8}{C{0.35in}}}
				\toprule
				\multirow{2}{4em}{} & \multirow{2}{4em}{\textbf{DoE}} & \multicolumn{4}{c}{\textbf{FB15k-237}} &
				\multicolumn{4}{c}{\textbf{WN18RR}} \\
				\cmidrule(lr){3-6} \cmidrule(lr){7-10}
				&  &   MRR   &   Hits@1   &   Hits@3   &   Hits@10   &   MRR   &   Hits@1   &   Hits@3   &   Hits@10 \\
				\midrule
				% \rowfont{\footnotesize}
				\multirow{4}{7em}{TGCN-Tucker}   &   32   &   .342   &   .254   &   .373   &   .516   &   .455   &   .419   &   .471   &   .521 \\
				&   64   &   .355   &   .262   &   .388   &   .537   &   .474   &   .434   &   .491   &   .545 \\
				&   100   &   .359   &  .266   &   .396   &   .542   &   .482   &   .441   &   .500   &   .560 \\
				% &   150   &   .339   &   .249   &   .370   &   .517   &   .482   &   .441   &   .500   &   .560 \\
				\bottomrule
			\end{tabular}
			}
	\label{tab:embeddingdim}
	% 	\vskip -0.1in
\end{table}

\subsection{The Effect of The Dimensionality of Embedding}
KGs commonly used in applications where there are numerous entities and relation types because of they represent abstract forms of information. Therefore, KGE models must poses the potential of scalability to large datasets. One essential feature to meet this scalability criteria is the dimensionality of embeddings. Therefore, in this experiment we validate the performance of TGCN when the dimensionality of embeddings are considerably low. Since the number of entity types is mostly the bottleneck of real-world applications, here, we change only the dimensionality of embeddings for entities $d_e$ while fixing it for relations as $d_r = 125$. 
 
Table \ref{tab:embeddingdim} shows the competitive performance of TGCN with low-domensional embeddings showing the potential of our model to be scaled and utilized for huge datasets. Especially when $d_e=64$ the model can still outperform most of the baselines except for TuckER, compared with Table 2, on both datasets while these methods use high-dimensional embeddings.

\subsection{The Effect of The Size of Random Subgraphs}
\label{sec:effectrandomsubgraph}
As we expected and shown in Figure \ref{fig:gs}, the size of random subgraphs $g_s$ considerably impacts the performance. This is due to our entity representation updating function (discussed in more detail in Section \ref{sec:subgraph}). Hyperparameters are tuned only with $g_s=50000$ on WN18RR, and fixed for other values of $g_s$.

Clearly, by increasing $g_s$, TGCN performance improves on both datasets. Nevertheless, there is only a slight performance increase in high values of $g_s$; thus, generalization might decrease after a peak in higher values of $g_s$. Concretely, by increasing $g_s$ each entity has broader access to its neighbors, providing it with more information. However, subgraph sampling loses its regularization effect (see Section \ref{sec:subgraph}) simultaneously, resulting in weaker performance.
\section{Conclusion}
We introduce TGCN, a general KG encoder, by incorporating Tucker decomposition in the aggregation function of R-GCN to efficiently integrate the information of relations and entities. Specifically, the projection matrices applied to the representations of neighbor entities depend on the relations. We compress our model using CP decomposition and discuss its considerable regularizing effect. Also, inspired by contrastive learning, we train our model using a cost function that tackle the scalability issue of the 1-N method for training on huge graphs. Our results show TGCN with embeddings of considerably lower dimensionality achieves superior performance to all the baselines on FB15k-237 and WN18RR.

\clearpage
\bibliographystyle{unsrtnat}
\bibliography{reference}

%%%%%%%
\clearpage
\appendix
\section{More Experiments}\label{appendix:experiments}
\subsection{CP Decomposition as Regularization Method}
To verify the claim that our model compression method can occasionally improve the performance of similar models as a regularizer, we use CP decomposition as an alternative for the original regularization methods in R-GCN (basis decomposition and block decomposition) \cite{schlichtkrull2018modeling}. Table \ref{tab:blockBasisCP} shows the performance of R-GCN using block decomposition, basis decomposition, and CP decomposition. These results are attained by implementing R-GCN using the original hyperparameters of the paper.

We can see that CP decomposition performs successfully as a regularization method and has decreased the values of \#NFP and \#EFP. Interestingly, this parameter reduction not only does not harm the model performance but also improves it.

\begin{table}[t!]
	\caption{KGC results of R-GCN on FB15k-237 and WN18RR using basis decomposition, block decomposition, and CP decomposition for regularization.}
	\vskip 0.1in
	\footnotesize
	\centerline{
			\begin{tabular}{L{0.3in}C{0.8in}C{0.2in}*{3}{C{0.35in}}}
				\toprule
				& Decomposition & DoE & MRR & \#NFP & \#EFP \\
				\midrule
				\multirow{3}{4em}{R-GCN} & Block & 500 & .238 & 2.32M & 7.39M  \\
				& Basis & 100 & .220 & 2.05M & 1.48M  \\
				& CP & 100 & .239 & 0.13M & 1.48M  \\
				\bottomrule
			\end{tabular}
	}
	\label{tab:blockBasisCP}
	% 	\vskip -0.1in
\end{table}

\section{Datasets}
\label{appendix:datasets}
We used FB15k-237 and WN18RR in our experiments which are subsets of FB15k and WN \cite{NIPS2013_1cecc7a7} and have the issue of information leakage of the training sets to test and validation sets. Statistics of the datasets are summarized in Table \ref{tab:datasetStatistics}.

\begin{table}[ht]
	\centering
	\caption{Statistics of the datasets used for evaluation. The number of triplets in training, test, and validation sets are shown. Also, $|\pazocal{V}|$ and $|\pazocal{R}|$ are the total number of entities and relations respectively.}
	\vskip 0.1in
	\small
    \begin{tabular}{L{0.8in}R{0.6in}R{0.6in}R{0.8in}R{0.8in}R{0.6in}}
		\toprule
		Datasets & \#Entities & \#Relations & Training & Validation & Test \\
		\midrule
		FB15k-237 & 14541  & 237 & 272115 & 17535 & 20466 \\
		WN18RR & 40943 & 11 & 86835 & 3034 & 3134 \\
		\bottomrule
	\end{tabular}
	\label{tab:datasetStatistics}
	\vskip -0.1in
\end{table}

\section{Experimental Details}
\label{appendix:experimentalSetups}

Our model is implemented in PyTorch \citep{paszke2019pytorch} and trained on single GPU core NVIDIA-L40. We used Adam \citep{DBLP:journals/corr/KingmaB14} to optimize our model and used $L_2$ norm of the relations and entities embedding matrices for regularization with a factor ($reg_f$) of $0.01$. The dimensionality of embeddings are fixed to $d_e=100$ and $d_r=125$ for the main experiments on both datasets. We chose the hyper-parameters based on the performance of our model on the validation set according to MRR using random search. We chose learning rate $lr$ out of $\{0.1, 0.01, 0.005, 0.001\}$, with a step decay of $0.95$ every $500$ iteration. The number of bases $n_b$ is fixed to $100$ on both datasets to keep \#ENFP low while attaining favorable performance. 

\begin{table*}[t!]
	\caption{Hyper-parameters for the results of the experiments.}
	\vskip 0.1in
	\centering
	\footnotesize
    \begin{tabular}{L{0.5in}C{0.4in}C{0.3in}C{0.5in}*{6}{R{0.2in}}}
		\toprule
		           Datasets & $g_s$ & $lr$ & Decoder & CP & $dr_i$ & $dr_{h1}$ & $dr_{h2}$ & $dr_o$ & $dr_d$ \\
		\midrule
		\multirow{4}{*}{FB15k-237} & \multirow{4}{*}{100000} & \multirow{4}{*}{0.005} & \multirow{2}{*}{Tucker} & No & 0.0 & 0.1 & 0.0 & 0.2 & 0.3 \\
            \cmidrule(lr){5-10}
            & & & & Yes & 0.2 & 0.1 & 0.1 & 0.2 & 0.3 \\
            \cmidrule(lr){4-10}
            & & & \multirow{2}{*}{Distmult} & No & 0.1 & 0.2 & 0.1 & 0.2 & \multirow{2}{*}{N/A} \\
            \cmidrule(lr){5-9}
		& & & & Yes & 0.1 & 0.1 & 0.1 & 0.2 & \\
            \midrule
		\multirow{4}{*}{WN18RR} & \multirow{4}{*}{50000} & \multirow{4}{*}{0.001} & \multirow{2}{*}{Tucker} & No & 0.0 & 0.0 & 0.0 & 0.3 & 0.3 \\
            \cmidrule(lr){5-10}
            & & & & Yes & 0.0 & 0.2 & 0.1 & 0.2 & 0.3 \\
            \cmidrule(lr){4-10}
            & & & \multirow{2}{*}{Distmult} & No & 0.0 & 0.1 & 0.0 & 0.2 & \multirow{2}{*}{N/A} \\
            \cmidrule(lr){5-9}
		& & & & Yes & 0.2 & 0.2 & 0.1 & 0.2 & \\
		\bottomrule
	\end{tabular}
	\label{tab:settingPrameters}
	\vskip -0.1in
\end{table*}

Next, all hyper-parameters are fixed except for dropout rates. We tuned dropout rates of encoder input ($d_i$), hidden states ($d_{h1}, d_{h2}$), output ($d_o$), and the decoder ($d_d$) in $\{0.0, 0.1, 0.2, 0.3\}$. We leave choosing higher $g_s, n_b$, and $d$ for future studies. Table \ref{tab:settingPrameters} contains all the settings in our experiments. 

\end{document}